\documentclass[letterpaper]{article} 
\usepackage{aaai23}  
\usepackage{times}  
\usepackage{helvet}  
\usepackage{courier}  
\usepackage[hyphens]{url}  
\usepackage{graphicx} 
\usepackage{natbib}  
\usepackage{caption} 
\usepackage{algorithm}
\usepackage{algorithmic}
\usepackage{array}
\usepackage{multirow}
\usepackage{multicol}
\usepackage{color}
\usepackage{caption}
\usepackage{subfigure}
\usepackage{amsfonts} 
\usepackage{booktabs}
\usepackage{amsmath}

\usepackage{newfloat}
\usepackage{listings}
\DeclareCaptionStyle{ruled}{labelfont=normalfont,labelsep=colon,strut=off} 
\floatstyle{ruled}
\newfloat{listing}{tb}{lst}{}
\floatname{listing}{Listing}
\title{FTM: A Frame-level Timeline Modeling Method for Temporal Graph Representation Learning}
\author{
    Qichen Ye\textsuperscript{\rm 1}\equalcontrib, Bowen Cao\textsuperscript{\rm 1}\equalcontrib, Weiyuan Xu\textsuperscript{\rm 1}, Yuexian Zou\textsuperscript{\rm 1,2}\thanks{Corresponding author.}
}
\affiliations{
    \textsuperscript{\rm 1}ADSPLAB, School of ECE, Peking University, Shenzhen, China, \\
    \textsuperscript{\rm 2}Peng Cheng Laboratory, Shenzhen, China

    \big\{yeeeqichen,zouyx\big\}@pku.edu.cn,
    \big\{cbw2021,xuwy\big\}@stu.pku.edu.cn\\
    
}

\usepackage{bibentry}

\begin{document}

\maketitle

\begin{abstract}
Learning representations for graph-structured data is essential for graph analytical tasks.
While remarkable progress has been made on static graphs, researches on temporal graphs are still in its beginning stage.
The bottleneck of the temporal graph representation learning approach is the neighborhood aggregation strategy, based on which graph attributes share and gather information explicitly.
Existing neighborhood aggregation strategies fail to capture either the short-term features or the long-term features of temporal graph attributes, leading to unsatisfactory model performance and even poor robustness and domain generality of the representation learning method.
To address this problem, we propose a Frame-level Timeline Modeling (\textbf{FTM}) method that helps to capture both short-term and long-term features and thus learns more informative representations on temporal graphs.
In particular, we present a novel link-based framing technique to preserve the short-term features
and then incorporate a timeline aggregator module to capture the intrinsic dynamics of graph evolution as long-term features.
Our method can be easily assembled with most temporal GNNs. 
Extensive experiments on common datasets show that our method brings great improvements to the capability, robustness, and domain generality of backbone methods in downstream tasks.
Our code can be found at \url{https://github.com/yeeeqichen/FTM}.
\end{abstract}

\section{Introduction}
\begin{figure}[t]
\centering
\includegraphics[width=0.47\textwidth]{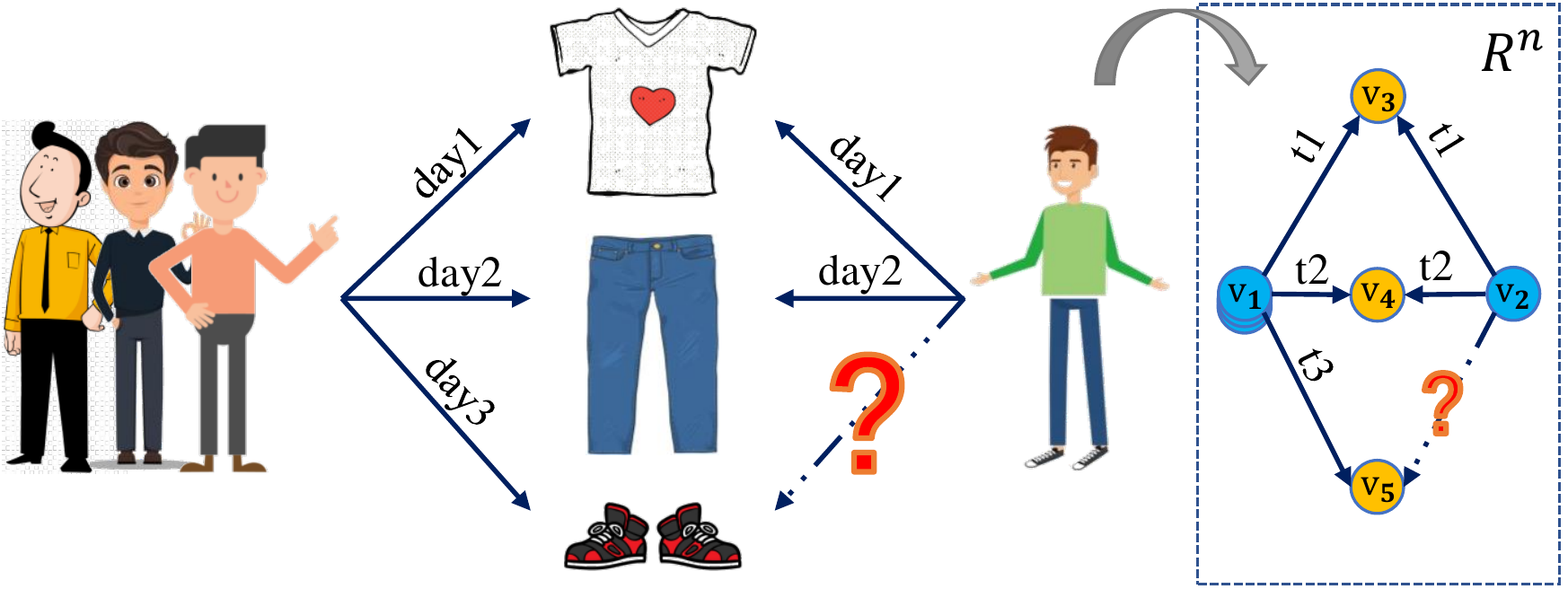} 
\caption{An example of temporal graph modeling. Given a model that has learnt the dynamics of a large number of users' shopping behaviors in high-dimensional space, what the man in green tends to buy in the future is predictable.}
\label{Fig.recom_system}
\end{figure}

Graph representation learning intends to transform nodes and links on the graph into lower-dimensional vector embeddings, which can be quite challenging due to the complex graph topological structures and node/link attributes. While approaches on \textbf{static graphs} have made breakthroughs and demonstrated distinguishable applicability in various fields \cite{graepel2010classic-user-pred2,he2014classic-user-ad,li2022joint, DGIF}, those on \textbf{temporal graphs} are just getting started. Modeling a temporal graph (which may evolve over time with the addition, deletion, and changing of its attributes) is a core problem in developing real-world industrial systems (\emph{e.g.}, social network, citation network, recommendation systems) where many data are time-dependent, and is much more difficult because of the temporal factors. Figure \ref{Fig.recom_system} gives an example of temporal graph modeling.

In learning representations on temporal graphs, a key point is the \textbf{neighborhood aggregation strategy}, which allows information passing and gathering among graph attributes, so that nodes learn their representations from their neighbors. For static graphs, directly linked nodes are neighbors to each other because they all appear in the one and only topology. In contrast, temporal graph attributes scatter sparsely across the timeline, leading to temporal-structure inconsistency. For any node in a temporal graph, a node connected to it is not necessarily a neighbor, for this node may appear a long time ago or disappear soon. Each node in a temporal graph may also have several temporal neighborhoods, posing a challenge for information aggregation.
Therefore, how to design the neighborhood aggregation strategy on temporal graphs remains an open question.

\begin{figure}[t]
\centering
\includegraphics[width=0.48\textwidth]{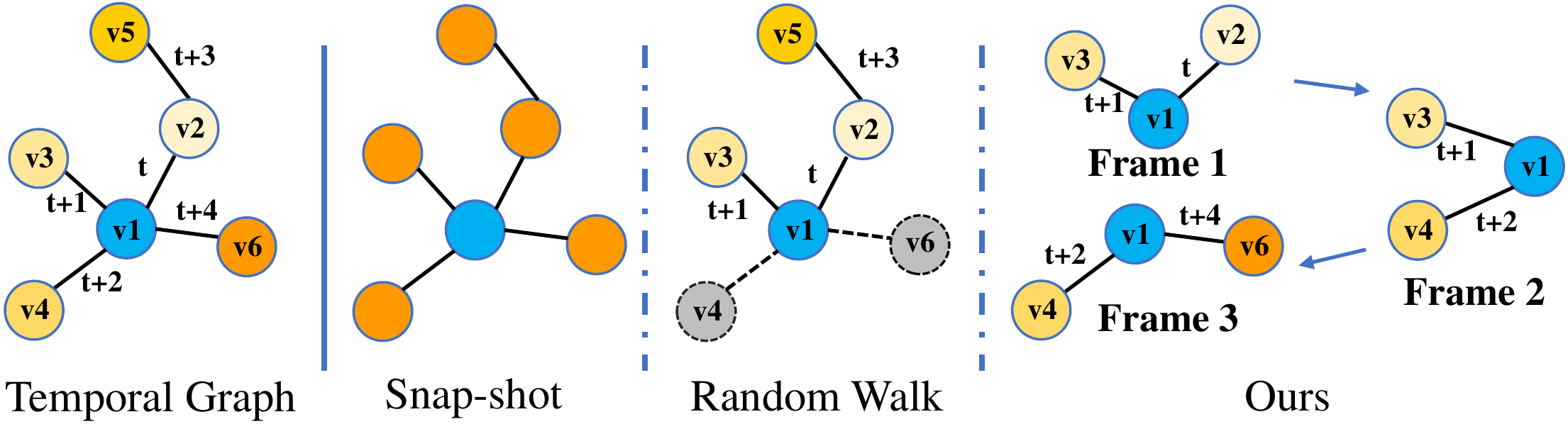} 
\caption{An example illustrating prior techniques and our link-based framing technique (where frame length is 2 and hop length is 1) for neighborhood construction.}
\label{Fig.framing_tech}
\end{figure}

Recent works introduce snapshot-based methods \cite{kumar2019jodie,pareja2020evolvegcn} and temporal random walk-based methods \cite{nguyen2018continuous-time,xu2020inductive} for neighborhood aggregation, but are often too simple to capture the evolution of temporal graphs over time.
The comparison of the above two methods and our method is shown in Figure \ref{Fig.framing_tech}.
In particular, snapshot-based methods equally slice the timeline into a sequence of snapshots, each of which contains nodes and links that occurred within its time span. This kind of method treats a snapshot as a static graph and fails to model the temporal properties within a snapshot, losing short-term features of graph attributes.
On the other hand, temporal random walk-based methods do not impose restrictions on the time range, but select temporal neighbors from the past according to a certain rule (most often randomly) and learn representations based on the neighborhood attributes and their time information.
However, the problem is that the randomly constructed temporal neighborhood cannot ensure a balance between short-term features and long-term features.

To develop a representation learning method on temporal graphs that adequately captures both short-term and long-term features, we propose a simple but effective \underline{\textbf{F}}rame-level \underline{\textbf{T}}imeline \underline{\textbf{M}}odeling method (\textbf{FTM} for short), at the heart of which is the innovation of the temporal neighborhood aggregation strategy: first, we refer to the concept of frame\footnote{A fundamental technique to decompose raw signal into multiple ranges according to frame length and hop length.} in signal processing, and put forward a novel method called \textbf{link-based framing technique}, where we separate most recent links into several frames (\emph{i.e.}, temporal neighborhoods) to emphasize short-term features; 
then, we extract frame features with a \textbf{frame aggregator}, which can be easily replaced by most GNN methods;
finally, we design a \textbf{timeline aggregator} for learning the intrinsic dynamics of successive frames across the timeline to capture long-term features.


We conduct experiments on several widely-used benchmarks in both transductive and inductive settings, and the results demonstrate the effectiveness of our proposed method. Moreover, the robustness and domain generality of baselines and our method are also evaluated through quantitative and qualitative experiments, which further suggest the insights of \textbf{FTM}.
Our main contributions are summarized as follows:

\begin{itemize}
\item We propose a simple but effective frame-level timeline modeling method for temporal graph representation learning, namely \textbf{FTM}, which makes contributions to the neighborhood aggregation strategy, and can be easily assembled with most GNN methods.
\item We conduct comprehensive experiments to show that models assembled with \textbf{FTM} achieve better performance on common benchmarks, and we further evaluate its effectiveness through quantitative and qualitative analyses.
\item We point out the robustness and domain generality issues of several state-of-the-art GNN-based temporal graph representation learning methods, and demonstrate that FTM could greatly alleviate these issues.
\end{itemize}

\section{Related Work}

Learning representations with GNNs has become a popular research area for graph modeling.
Earlier works explore learning representations of topological structures \cite{kipf2016GCN,grover2016node2vec}, extending GNN to inductive learning \cite{hamilton2017inductive}, and integrating attention mechanisms \cite{velivckovic2017graph}.
In all these works, however, the time information of graph attributes are discarded.

Recent approaches take advantage of the temporal property. Certain approaches learn to access time-aware knowledge by equally slicing the timeline into a sequence of \textbf{snapshots}
\cite{trivedi2019dyrep,singer2019tNodeembed}. They aggregate the topological features in a snapshot and combine time-dependent features with sequence-modeling techniques to learn temporal graph embeddings.
However, they ignore the sequential nature of nodes and links within the same snapshot, losing short-term features that can guide learning. Meanwhile, the amount of nodes and links within each snapshot is inconsistent, leading to great data biases in learning topological features.

More recently, TGAT \cite{xu2020inductive} leverages a time encoding function to learn time-aware representations in continuous time. TGN \cite{rossi2020temporal}, as a variant of TGAT, integrates a memory module to keep track of the evolution of node-level features. These methods make progress in capturing short-term features since the time encoding makes it possible to model the temporal properties of a neighborhood.
However, in most cases, they randomly sample neighbors from the past to form a temporal neighborhood for a target node, which means that they cannot ensure a balance between short-term features and long-term features.

Our work adopt the idea of time encoding, but make contributions to the way that temporal neighborhoods are constructed and information is aggregated, so that the model learn more informative representations.

\begin{figure*}[t]
\centering
\includegraphics[width=1.0\textwidth]{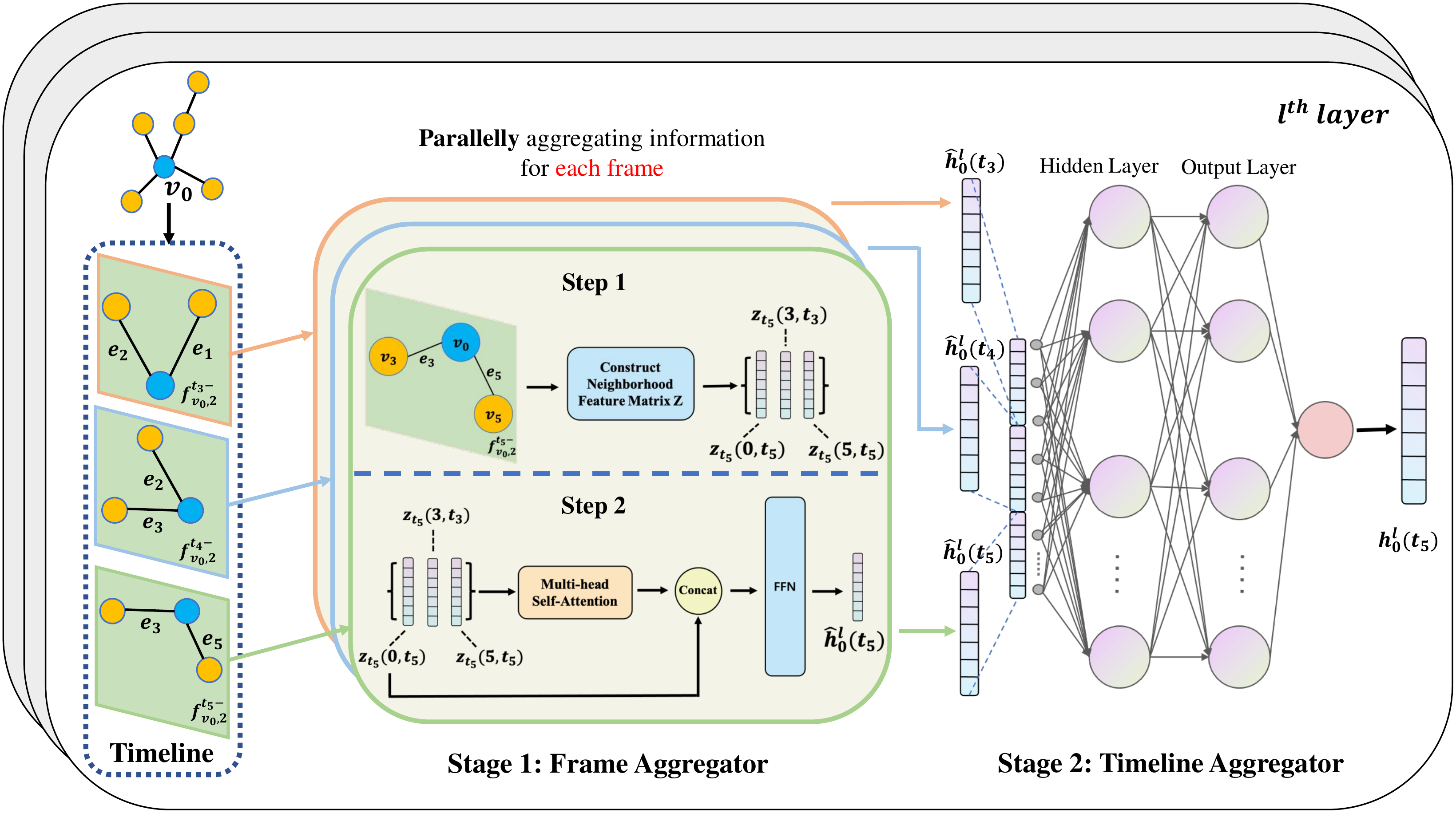} 
\caption{The architecture of the model assembling FTM with a backbone network. Assuming that our goal is to compute node $v_0$'s representation at timestamp $t_5$, we first construct a timeline consists of 3 frames 
$\bigl\{ f^{t_3-}_{v_0,2}, f^{t_4-}_{v_0,2}, f^{t_5-}_{v_0,2} \bigr\}$ 
as each layer's input. At each layer - \textbf{Stage 1}, we compute each frame's representation
$\hat{{\textbf{h}}}^l_0(t_j)$ 
in parallel through the backbone network (works as the frame aggregator). \textbf{Stage 2}, we aggregate all frames' representations to get the node representation via the timeline aggregator.
}
\label{Fig.model structure}
\end{figure*}

\section{Proposed Method: \textbf{FTM}}


\subsection{Problem Formalization}

Graph representation learning aims to obtain node or link representations based on their own properties and their interactions with neighbors.
Let $E^{T-} = \bigl\{ e_{i,j,t} | 1 \leq i, j \leq n, 0 \leq t < T \bigr\}$ and $V^{T-} =\bigl\{ v_s | s=1 \dots n \bigr\} $ denote the set of links and the set of nodes observed before time $T$, respectively, where $n$ is the amount of nodes, $v_s$ is the $s$-th node ($s$ is only used to distinguish nodes), and $e_{i,j,t}$ is an link between $v_i$ and $v_j$ emerged at time $t \in \mathbb{R}^{+}$. 
Let $E_s^{T-}= \bigl\{ e_{s,i,t} | 1 \leq i \leq n, 0 \leq t < T \bigr\} \cup \bigl\{ e_{j,s,t} | 1 \leq j \leq n, 0 \leq t < T \bigr\}$ denotes the subset of $E^{T-}$ containing links that link to node $v_s$ and satisfy the time constraint (we mainly consider undirected graphs, where the two parts of $E_s^{T-}$ are equivalent).
Supposing that $G^{T-}=(V^{T-}, E^{T-})$ denotes the final state of a temporal graph before time $T$, learning representations on it is mainly to obtain the node and link representations at time $t$ based on $G^{T-}$. 

\subsection{Input Representation}
\label{input_representation}
Graph attributes can be recorded in various ways. For instance, online reviews are in text format, and citations are in triplet format. We encode text with BERT-base \cite{devlin2018bert}, and other records with TransE \cite{2013Translating}, to initialize node and link features.
Then, we split links into frames, and feed the features of successive frames into FTM.

\noindent\textbf{Link-based Framing Technique.} \label{framing}
The process of splitting links into temporal frames is controlled by two parameters:

\begin{itemize}
\item[-]
\textbf{Frame length} defines how many links are included in a frame. 
For example, at timestamp $t$, to construct a frame of length $k$ for node $v_s$, we take the most recent $k$ links from $E_s^{t-}$ to form this frame and denote it as $f_{s,k}^{t-}$.
\item[-]
\textbf{Hop length} defines how many links to skip when taking the next frame. In practise, we set it to be $\frac{frame\ length}{2}$ (which is empirically the best and is also a convention in signal processing) to stabilize the training process. An example is provided in Figure \ref{Fig.framing_tech}.
\end{itemize}

\subsection{Frame-level Timeline Modeling\label{FTM method}}

The main idea of FTM is to preserve both the short-term and long-term features of graph attributes through a \textbf{frame aggregator} and a \textbf{timeline aggregator}. The role of the frame aggregator is to model each neighborhood that generated by the link-based framing technique, so \textbf{it can be replaced by most GNN methods}.
For example, the overall framework of the model assembling FTM with TGAT \cite{xu2020inductive}, \emph{i.e.}, taking TGAT as the frame aggregator, is shown in Figure \ref{Fig.model structure}.
Since TGAT is composed of a stack of identical layers (with shared parameters), the calculation process of each layer is similar. Assuming that we want node $v_i$'s embedding at timestamp $t$, the calculation process in layer $l$ can be described as the following two parts:

\vspace*{1mm}
\noindent\textbf{Temporally Attentive Frame Aggregator.}
While TGAT randomly samples links from the past to form temporal neighborhoods, we integrate \textbf{$k$ most recent} links to construct a frame in order to preserve short-term features. Meanwhile, the reason we add links by number rather than by time (as snapshot-based methods) is to guide the model to learn the common evolution of links, instead of time-interval-related knowledge.
Given a frame $f^{t-}_{i, k}$ of $v_i$ that contains links $\bigl\{ e_{i,j_1, t_1},\dots,e_{i, j_k, t_k}\bigr\}$, we obtain a temporal neighborhood feature matrix $\textbf{Z}(t)$ as:
\begin{equation}
    \textbf{Z}(t) = \left[\textbf{z}_{t}(i,t),\textbf{z}_{t}(j_1,t_1),\dots,\textbf{z}_{t}(j_k,t_k)\right],
\end{equation}
\begin{equation}
    \textbf{z}_{t}(j_k, t_k) = \left[\textbf{h}^{(l -1)}_{j_k}(t_k)\ ||\ \varphi(t - t_k)\ ||\ \textbf{e}_{j_k}\right],
    \label{equa: z_t()}
\end{equation}
where $\textbf{h}^{(l - 1)}_{j_k}(t_k)$ is the previous layer's output for $v_{j_k}$, $\varphi(\cdot)$ is a time encoding function, $\textbf{e}_{j_k}$ is the feature vector of $e_{i, j_k, t_k}$, and $\textbf{z}_{t}(j_k, t_k)$ maps the information of $v_{j_k}$ into a time-aware representation.
Then, we attentively aggregate $Z_t$ with the multi-head self-attention mechanism:
\begin{equation}
    \textbf{q}^{r}(t) = [\textbf{Z}(t)]_0 \textbf{W}^{r}_{Q}, 
\end{equation}
\begin{equation}
     \textbf{K}^{r}(t) = [\textbf{Z}(t)]_{1:N}\textbf{W}^{r}_K,
    \ \textbf{V}^{r}(t) = [\textbf{Z}(t)]_{1:N}\textbf{W}^{r}_V
\end{equation}
\begin{equation}
     \alpha^{r}_j = \frac{\exp\left({ {\textbf{q}^{r}}^{\top}\textbf{K}^{r}_j}\right)}{\Sigma_q \exp\left({{\textbf{q}^{r}}^{\top}\textbf{K}^{r}_q}\right)},
     \ \tilde{\textbf{h}}^{l, r}_{i}(t) = {\sum}_j \alpha^{r}_j \textbf{v}^{r}_j,
\end{equation}
where $\textbf{W}^{r}_Q, \textbf{W}^{r}_K, \textbf{W}^{r}_V$ are query, key and value matrix, respectively, $\alpha^{r}_j$ denotes the attention weight, and $\tilde{\textbf{h}}^{l, r}_{i}(t)$ is the output of the $r$-th attention head.
Assuming that we have $n_h$ attention heads, the frame representation $\hat{\textbf{h}}^l_i(t)$ will be:
\begin{equation}
    \hat{\textbf{h}}^{l}_i(t) = \operatorname{ReLU}(\textbf{y}\textbf{W}_0 + \textbf{b}_0) \textbf{W}_1 + \textbf{b}_1 ,
\end{equation}
\begin{equation}
    \textbf{y} = \left[\textbf{z}_t(i, t) || \tilde{\textbf{h}}^{l, 1}_i(t)|| \dots ||\tilde{\textbf{h}}^{l, n_h}_i(t)\right] ,
\end{equation}
where $\textbf{W}_0$, $\textbf{W}_1$ are weights and $\textbf{b}_0, \textbf{b}_1$ are biases.



\vspace*{1mm}
\noindent\textbf{Attentively Frame-level Timeline Aggregator.} 
In the prior part, we get the representation $\hat{\textbf{h}}^l_i(t)$ of frame $f^{t-}_{i, k}$. Now, we consider how to aggregate the information of multiple frames. Empirically, we set the hop length to half of the frame length to retain redundant information between frames. By doing so, \emph{(i)} short-term features are further highlighted; and \emph{(ii)} framing serves as a \textbf{scrubbing technique} because irregular links (with abnormal time interval/content) will not play a leading role and the commonalities in the evolution of links will be emphasized.

Let $F^{t-}_{i,k} = \bigl\{ f^{t_j-}_{i,k}| 1 \leq j \leq n, t_n = t\bigr\}$ denotes a set of frames of node $v_i$, in which the timestamps satisfy: 
\begin{equation}
t_{j-1} = \mathbb{T}_{\frac{k}{2}}(E^{t_j-}_i), 2 \leq j \leq n
\label{equa: frame_timestamp}
\end{equation}
where $n$ is the size of this set, $\mathbb{T}_{\frac{k}{2}}$ maps a set of links to its $\frac{k}{2}$-th (\emph{i.e.}, half of the frame length) recent element's timestamp. We call this set a $n$-length \textbf{timeline} of node $v_i$ at timestamp $t$, and we get the final node representation $\textbf{h}^l_{i}(t)$ as:
\begin{equation}
    \textbf{h}^l_{i}(t) = \left[\hat{\textbf{h}}^l_{i}(t_1) || \dots|| \hat{\textbf{h}}^l_{i}(t_n)\right]^{T}\textbf{W}_2 + \textbf{b}_2,
\label{equa: h_final}
\end{equation}
where $\textbf{W}_2$ and $\textbf{b}$ are weights and bias. Here we take $1$-layer MLP as an example for simplicity, but it could be effortlessly extended to RNN-based or attention-based methods, \emph{etc.} 

$\textbf{h}^l_{i}(t)$ generated by the last layer is just what we want - node $v_i$’s embedding at timestamp $t$, $\textbf{h}_{i}(t)$.

\vspace{1mm}
\noindent\textbf{Learning \& Inference.} 
Since the temporal information is mostly reflected in the time-sensitive interactions among nodes, we choose to use the future link prediction setup for training.
The goal of future link prediction is to predict the probability that an link will exist between a target node $v_i$ and another node $v_j$ at a specific future time, \emph{i.e.}, given the set of previous links of $v_i$, we compute the probability of a future link $e_{i,j,t_{i,j}}$ between $v_i$ and $v_j$.
To train the model, we sample a set of negative links ($\neq e_{i,j,t_{i,j}}$) and optimize the per-node objective:
\begin{equation}
L=\sum\limits_{v_i, v_j, t_{i,j}} \operatorname{Pos}\left(i, j, t_{i,j}\right) + Q \cdot  E_{v_q \sim P} \operatorname{Neg}\left(i, q, t_{i,j}\right)
\end{equation}
where $P$ is the negative link sampling distribution, $Q$ denotes the negative sampling size, $\operatorname{Pos}(\cdot,\cdot,\cdot)$ and $\operatorname{Neg}(\cdot,\cdot,\cdot)$ denote the positive and negative scoring functions:
\begin{equation}
\operatorname{Pos}\left(i, j,t_{i,j}\right) = -\log\left(\sigma\left(-\textbf{h}_i(t_{i,j})^{\top}\textbf{h}_j(t_{i,j})\right)\right)
\end{equation}
\begin{equation}
\operatorname{Neg}\left(i, q,t_{i,j}\right) =  -\log\left(\sigma\left(\textbf{h}_i(t_{i,j})^{\top}\textbf{h}_q(t_{i,j})\right)\right)
\end{equation}
where $\sigma(\cdot)$ is an activation function, $\textbf{h}_i(t)$ is the representation of node $v_i$ at timestamp $t$. For inference, the output of $\operatorname{Pos}(i, j, t_{i,j})$ is used as the logits.

\begin{table}[t]
\scriptsize
\centering
\setlength{\tabcolsep}{5pt}
\begin{tabular}{ccc}
\hline
\textbf{Dataset} & \textbf{Node} & \textbf{Link} \\
\hline
\specialrule{0em}{0.5pt}{0pt}
Reddit \cite{kumar2019jodie} & 11,000 & 672,000 \\
\specialrule{0em}{0.5pt}{0pt}
Wikipedia \cite{kumar2019jodie} & 9,000 & 157,000 \\
\specialrule{0em}{0.5pt}{0pt}
Icews14 \cite{GarcaDurn2018LearningSE} & 7,000 & 91,000 \\
\specialrule{0em}{0.5pt}{0pt}
Icews05-15 \cite{GarcaDurn2018LearningSE} & 10,000 & 461,000 \\
\specialrule{0em}{0.5pt}{0pt}
Bitcoin-otc \cite{kumar2016edge} & 6,000 & 36,000 \\
\specialrule{0em}{0.5pt}{0pt}
Bitcoin-alpha \cite{kumar2016edge} & 4,000 & 24,000 \\
\specialrule{0em}{0.5pt}{0pt}
Mooc \cite{kumar2019jodie} & 7,000 & 412,000 \\
\specialrule{0em}{0.5pt}{0pt}
\hline
\end{tabular}
\caption{\label{scale of datasets}
The node and link statics for each dataset.
}
\end{table}

\begin{table*}[t]
\scriptsize
\centering
\setlength{\tabcolsep}{15pt}
\begin{tabular}{ccccc}
\toprule
\multirow{2}{*}{\textbf{Model}} & \multicolumn{2}{c}{\textbf{Reddit}} & \multicolumn{2}{c}{\textbf{Wikipedia}} \\
\cmidrule(lr){2-3}
\cmidrule(lr){4-5}
& \multicolumn{1}{c}{Transductive} & \multicolumn{1}{c}{Inductive} & \multicolumn{1}{c}{Transductive} & \multicolumn{1}{c}{Inductive} \\
\hline
GAE \cite{kipf2016GAE} & 93.23 & - & 91.44 & - \\
VAGE \cite{kipf2016GAE} & 92.92 & - & 91.34 & - \\
DeepWalk \cite{perozzi2014deepwalk} & 83.10 & - & 90.71 & - \\
Node2vec \cite{grover2016node2vec} & 84.56 & - & 91.48 & - \\
CTDNE \cite{nguyen2018continuous-time} & 91.41 & - & 92.17 & - \\
DyRep \cite{trivedi2019dyrep} & 98.25 & 96.11 & 94.76 & 92.11 \\
Jodie \cite{kumar2019jodie} & 97.02 & 94.46 & 92.75 & 93.13 \\
\hline
GraphSAGE \cite{hamilton2017inductive} & 97.20 & 94.68 & 91.09 & 86.08 \\
\textbf{w/ FTM} & 98.01$\uparrow$ & 96.28$\uparrow$ & 92.91$\uparrow$ & 91.93$\uparrow$ \\
\cline{2-5}
GAT \cite{velivckovic2017graph} & 97.33 & 95.37 & 94.73 & 91.27 \\
\textbf{w/ FTM} & 98.21$\uparrow$ & 96.75$\uparrow$ & 95.03$\uparrow$ & 93.54$\uparrow$ \\
\cline{2-5}
TGAT \cite{xu2020inductive} & 98.27 & 96.73 & 95.13 & 93.97 \\
\textbf{w/ FTM} & 98.41$\uparrow$ & 96.82$\uparrow$ & 97.82$\uparrow$ & 97.14$\uparrow$ \\
\cline{2-5}
TGN \cite{rossi2020temporal} & 98.78 & 97.77 & 98.28 & 97.69  \\
\textbf{w/ FTM} & \textbf{98.88}$\uparrow$ & \textbf{97.96}$\uparrow$ & \textbf{98.82}$\uparrow$ & \textbf{98.33}$\uparrow$ \\
\cline{2-5}
\color{red}{Average Gain} & \color{red}{0.48} & \color{red}{0.82} & \color{red}{1.34} & \color{red}{2.98} \\
\bottomrule
\end{tabular}
\caption{\label{result-trans-induc}
AP(\%) for future link prediction tasks. $\uparrow$ means that FTM brings an improvement to the baseline method. The best results in each column are highlighted in \textbf{bold} font. '-' denotes incapability.}
\end{table*}

\begin{table*}[t]
\centering
\scriptsize
\setlength{\tabcolsep}{10.8pt}
\begin{tabular}{cccccccc}
\toprule
\multirow{2}{*}{\textbf{Model}} & \multicolumn{7}{c}{\textbf{Attack Intensity(\%)}} \\
\cline{2-8}
& \textbf{0} & \textbf{1} & \textbf{10} & \textbf{20} & \textbf{30} & \textbf{40} & \textbf{50} \\
\hline
GraphSAGE (\citeyear{hamilton2017inductive}) & 85.46(+1.58) & 62.24(+19.52) & 34.11(+28.14) & 51.78(+3.56) & 45.89(+5.85) & 40.81(+3.04) & 48.84(+9.15) \\
\cline{2-8}
GAT (\citeyear{velivckovic2017graph}) & 83.75(+4.31) & 79.16(+5.11) & 49.56(+17.66) & 40.47(+19.32) & 41.89(+16.11) & 36.70(+9.39) & 41.38(+15.06) \\
\cline{2-8}
TGAT (\citeyear{xu2020inductive}) & 87.36(+1.37) & 87.67(+1.16) & 59.99(+26.83) & 56.59(+29.85) & 47.39(+38.39) & 34.61(+51.56) & 38.57(+47.29) \\
\cline{2-8}
TGN (\citeyear{rossi2020temporal}) & 88.19(+1.82) & 86.68(+1.99) & 80.80(+3.18) & 81.93(+2.63) & 81.81(+4.96) & 83.39(+2.09) & 83.17(+2.40)  \\
\hline
\color{red}{Average Gain} & \color{red}{2.27} & \color{red}{6.95} & \color{red}{18.95} & \color{red}{13.84} & \color{red}{16.33} & \color{red}{16.52} & \color{red}{18.48} \\
\bottomrule
\end{tabular}
\caption{\label{robust-attack-2}
AUC (\%) for node classification tasks on Wikipedia. Attack intensity controls the ratio of (the norm of) the added noise to (the maximum norm of) the link features in the dataset. \textbf{x(+y)} indicates that the baseline method achieves \textbf{x\%} in AUC, and FTM brings an improvement of \textbf{y\%} to it, \emph{i.e.}, the model assembling FTM with this method achieves \textbf{x+y\%}.}
\end{table*}

\section{Experimental Setups\label{experiments}}
We evaluate our method against strong baselines (adapted to temporal settings when possible). Note that \textbf{assembling FTM with a baseline method} means that we take the baseline method as the frame aggregator of FTM.

\subsection{Tasks and Metrics}

We perform future link prediction to evaluate the quality of the generated graph representations. We use average precision (AP) as the evaluation metric and consider this task in two settings: \emph{(i)} \textbf{Transductive Task.}
We predict future links among nodes that have been observed during training. \emph{(ii)} \textbf{Inductive Task.} 
We perform future link prediction among nodes that have not been observed in the training phase.

\subsection{Datasets\label{datasets}}

We choose seven datasets that contain time-sensitive node interactions:
\noindent\textbf{Reddit}\footnote{http://snap.stanford.edu/jodie/reddit.csv} is created from posts between active users and subreddits, where users and subreddits are nodes, and posts are links.
\noindent\textbf{Wikipedia}\footnote{http://snap.stanford.edu/jodie/wikipedia.csv} is created by taking top edited pages in Wikipedia and active users as nodes, and the corresponding edits as links.
\noindent\textbf{Icews14}\footnote{https://github.com/nle-ml/mmkb}, \textbf{Icews05-15}\footnote{https://github.com/nle-ml/mmkb} contain political events and the corresponding timestamps. 
All nodes are real-world entities (e.g. countries) and links are event types.
\noindent\textbf{Bitcoin-otc}\footnote{https://snap.stanford.edu/data/soc-sign-bitcoinotc.csv.gz}, \textbf{Bitcoin-alpha}\footnote{https://snap.stanford.edu/data/soc-sign-bitcoinalpha.csv.gz} are who-trusts-whom networks of people who trade with Bitcoin, where nodes are people and links are the credit evaluation.
\noindent\textbf{Mooc}\footnote{https://snap.stanford.edu/data/act-mooc.tar.gz} dataset contains user actions on a popular MOOC platform, where nodes represent users and course activities, and links represent user actions.
Dataset scales are listed in Table \ref{scale of datasets}.


\subsection{Baselines}

\noindent\textbf{GAE}, \textbf{VAGE} \cite{kipf2016GAE}, \textbf{DeepWalk} \cite{perozzi2014deepwalk} and \textbf{Node2vec} \cite{grover2016node2vec} are models for static graphs.

\noindent\textbf{CTDNE}, \textbf{DyRep}, \textbf{Jodie}, \textbf{GraphSAGE}, \textbf{GAT}, \textbf{TGAT} and \textbf{TGN} are baselines for temporal graphs.
We do not ensemble FTM with CTDNE, DyRep and Jodie due to the conflicting schemes\footnote{These methods have their own custom temporal neighborhood construction strategies. If we apply our action-based framing technique to these methods, we are only assembling FTM with their feature extraction modules.}.
For other methods, we test the original version and the FTM-assembled version. 
There may be slight differences between our implementation and others, but it is fair for comparison.

\begin{figure*}[t]
\centering
	\subfigure[Classify users of Wiki]{
	\begin{minipage}{0.45\linewidth}
		\centering
		\includegraphics[width=5.69cm]{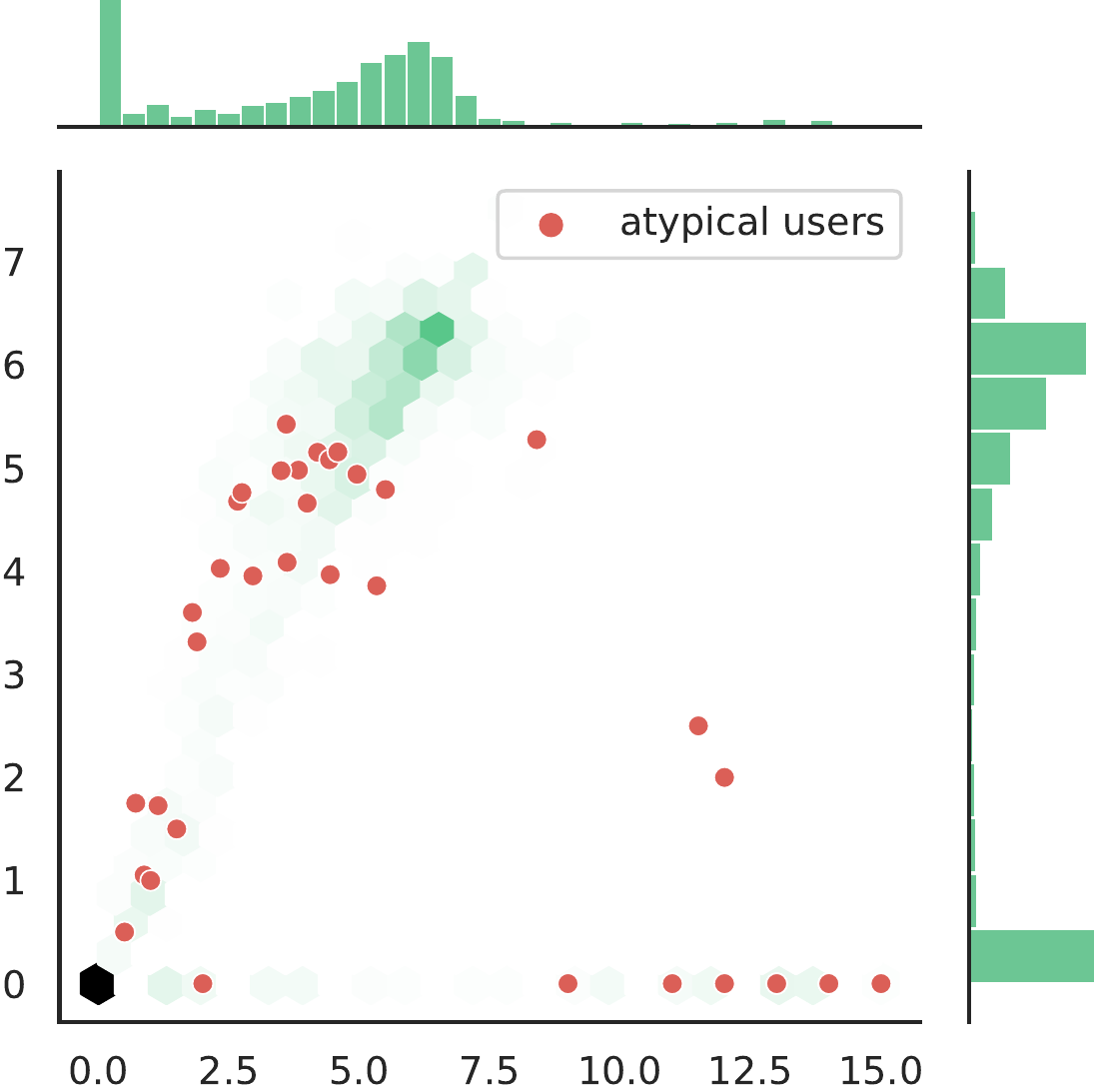}
	\label{node classify on wiki}
	\end{minipage}}
	\subfigure[Embedding analysis]{
	\begin{minipage}{0.45\linewidth}
		\centering
		\includegraphics[width=5.49cm]{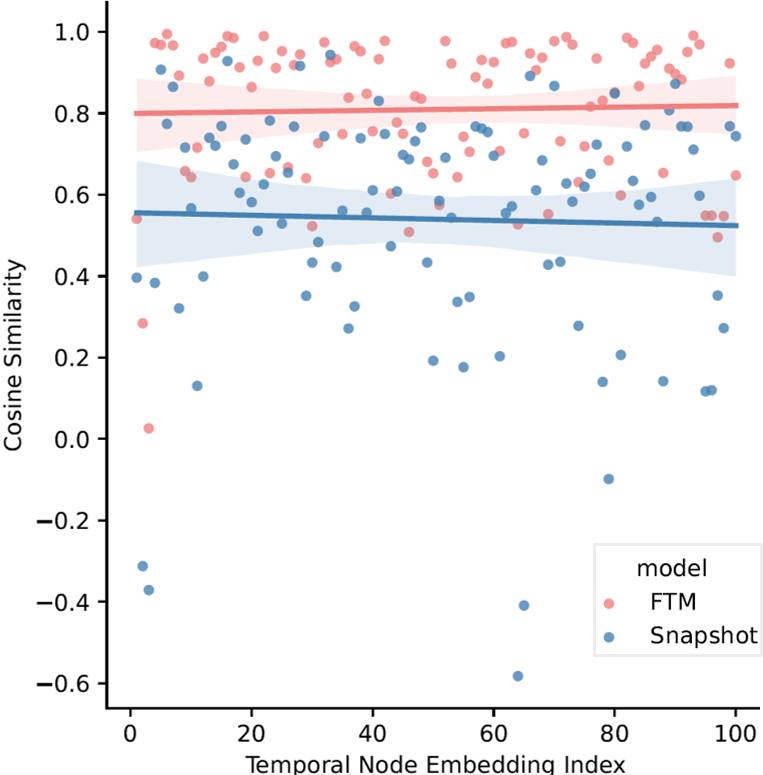}
	\label{Fig.frame_snapshot_cos_sim}
	\end{minipage}}
	\caption{(a) x-axis/y-axis represents the average/standard deviation of the time intervals of a user's actions. The green parts denotes user distribution. The darker the color, the greater the number of users. Red points denote atypical users that have misled TGAT but are correctly classified by TGAT+FTM. (b) The cosine similarity of successive temporal node embeddings generated by TGAT+FTM and TGAT+Snapshot, respectively. 
    The consistency of the embeddings generated by TGAT+FTM proves that FTM helps to learn stable temporal representations.}
\end{figure*}

\begin{table*}[t]
\scriptsize
\centering
\setlength{\tabcolsep}{15pt}
\begin{tabular}{ccccccc}
\toprule
\multirow{1}{*}{\textbf{Training}} & \multirow{2}{*}{\textbf{Model}} & \multicolumn{5}{c}{\textbf{Test Dataset}} \\
\cline{3-7}
\textbf{Dataset} & & \textbf{Icews14} & \textbf{Icews05-15} & \textbf{Bitcoin-otc}& \textbf{Bitcoin-alpha} &
\textbf{Mooc} \\
\hline
\multirow{5}{*}{\textbf{Reddit}} & GraphSAGE (\citeyear{hamilton2017inductive}) & 46.89(+35.32) & 61.48(+23.08) & 70.36(+7.59) & 54.44(+16.09) & 49.86(+3.38) \\
\cline{3-7}
& GAT (\citeyear{velivckovic2017graph}) & 63.45(+24.32) & 64.44(+20.81) & 70.66(+6.30) & 61.35(+9.49) & 47.28(+7.25) \\
\cline{3-7}
& TGAT (\citeyear{xu2020inductive}) & 76.29(+9.82) & 72.80(+15.47) & 70.19(+10.81) & 65.46(+8.11) & 57.01(+16.98) \\
\cline{3-7}
& TGN (\citeyear{rossi2020temporal}) & 68.63(+12.20) & 70.57(+15.72) & 72.86(+6.48) & 64.55(+6.04) & 67.23(+2.48)  \\
\cline{3-7}
& \color{red}{Average Gain} & \color{red}{20.42} & \color{red}{18.77} & \color{red}{7.80} & \color{red}{9.93} & \color{red}{7.52} \\
\hline
\multirow{5}{*}{\textbf{Wikipedia}} & GraphSAGE (\citeyear{hamilton2017inductive}) & 71.88(+7.59) & 77.49(+3.46) & 58.88(+12.44) & 53.81(+18.16) & 49.11(+4.85)  \\
\cline{3-7}
& GAT (\citeyear{velivckovic2017graph}) & 67.19(+12.29) & 69.32(+15.20) & 67.20(+0.34) & 61.48(+6.71) & 49.42(+7.19) \\
\cline{3-7}
& TGAT (\citeyear{xu2020inductive}) & 80.27(+6.94) & 82.03(+10.82) & 71.38(+12.16) & 71.01(+2.18) & 53.98(+22.54) \\
\cline{3-7}
& TGN (\citeyear{rossi2020temporal}) & 66.40(+15.73) & 67.77(+16.36) & 83.76(+0.41) & 64.69(+7.29) & 73.20(+1.86) \\
\cline{3-7}
& \color{red}{Average Gain} & \color{red}{10.64} & \color{red}{11.46} & \color{red}{6.34} & \color{red}{8.59} & \color{red}{9.11} \\
\bottomrule
\end{tabular}
\caption{\label{result-generalization-2}
AP (\%) of future link prediction tasks.
\textbf{x(+y)} indicates that the baseline method achieves \textbf{x\%} in AP, and FTM brings an improvement of \textbf{y\%} to it, \emph{i.e.}, the model assembling FTM with this method achieves \textbf{x+y\%}.}
\end{table*}

\begin{table*}[t]
\centering
\scriptsize
\setlength{\tabcolsep}{4.2pt}
\begin{tabular}{c|cccccccc|cccccccc}
\toprule
\multirow{3}{*}{\textbf{Model}} & \multicolumn{8}{c}{\textbf{Neighborhood Scale}} & \multicolumn{8}{c}{\textbf{Percentage of Training Data}} \\
& \multicolumn{4}{c}{\textbf{Inductive}} & \multicolumn{4}{c}{\textbf{Generalization}} & \multicolumn{4}{c}{\textbf{Inductive}} & \multicolumn{4}{c}{\textbf{Generalization}} \\
\cline{2-17}
& \textbf{S} & \textbf{M} & \textbf{L} & \textbf{XL} & \textbf{S} & \textbf{M} & \textbf{L} & \textbf{XL} & \textbf{1\%} & \textbf{5\%} & \textbf{10\%} & \textbf{50\%} & \textbf{1\%} & \textbf{5\%} & \textbf{10\%} & \textbf{50\%} \\
\hline
GraphSAGE & 86.31 & 88.96 & 94.19 & 94.68 & 70.87 & 70.83 & 78.74 & 83.59 & 65.31 & 85.39 & 91.17 & 95.64 & 57.99 & 62.79 & 73.04 & 80.15 \\
\textbf{w/ FTM} & 92.24$\uparrow$ & 92.31$\uparrow$ & 95.53$\uparrow$ & 96.28$\uparrow$ & 79.37$\uparrow$ & 77.26$\uparrow$ & 86.30$\uparrow$ & 86.53$\uparrow$ & 70.40$\uparrow$ & 87.58$\uparrow$ & 91.95$\uparrow$ & 96.65$\uparrow$ & 61.98$\uparrow$ & 74.92$\uparrow$ & 82.71$\uparrow$ & 85.34$\uparrow$ \\
\cline{2-17}
GAT & 91.11 & 93.15 & 95.56 & 95.37 & 69.88 & 74.96 & 83.76 & 85.84 & 68.99 & 90.81 & 93.13 & 95.10 & 59.53 & 76.44 & 79.70 & 85.80 \\
\textbf{w/ FTM} & 91.85$\uparrow$ & 93.40$\uparrow$ & 95.84$\uparrow$ & 96.75$\uparrow$ & 82.38$\uparrow$ & 81.75$\uparrow$ & 86.20$\uparrow$ & 88.97$\uparrow$ & 73.13$\uparrow$ & 91.02$\uparrow$ & 93.70$\uparrow$ & 96.68$\uparrow$ & 68.45$\uparrow$ & 81.99$\uparrow$ & 85.91$\uparrow$ & 90.30$\uparrow$ \\
\cline{2-17}
TGAT & 91.12 & 92.63 & 95.95 & 96.73 & 69.22 & 71.76 & 85.64 & 87.34 & 65.65 & 88.92 & 92.67 & 96.25 & 74.51 & 77.16 & 81.27 & 86.38 \\
\textbf{w/ FTM} & 94.08$\uparrow$ & 94.32$\uparrow$ & 97.26$\uparrow$ & 96.82$\uparrow$ & 91.08$\uparrow$ & 89.52$\uparrow$ & 95.82$\uparrow$ & 91.06$\uparrow$ & 80.76$\uparrow$ & 92.32$\uparrow$ & 93.45$\uparrow$ & 96.25 & 81.84$\uparrow$ & 87.88$\uparrow$ & 87.22$\uparrow$ & 88.53$\uparrow$ \\
\hline
\color{red}{Average Gain} & \color{red}{3.21} & \color{red}{1.76} & \color{red}{0.98} & \color{red}{1.02} & \color{red}{14.29} & \color{red}{10.33} & \color{red}{6.73} & \color{red}{3.26} & \color{red}{8.10} & \color{red}{1.67} & \color{red}{0.64} & \color{red}{0.69} & \color{red}{5.00} & \color{red}{8.25} & \color{red}{5.69} & \color{red}{2.87} \\
\bottomrule
\end{tabular}
\caption{\label{case-neighborhoodsize}
Case studies on (1) neighborhood scale, where neighborhood scale expands from S to XL; and (2) the percentage of training data, where models are trained on limited training data of Reddit, \emph{e.g.}, 1\% means models are trained/validated on one-percent of the original training/validation data. We do not take TGN into consideration, because the way TGN updates node-wise memory has little to do with the neighborhood scale and the percentage of training data. We report AP(\%) of future link prediction on Reddit (inductive; generalize from Wiki). }
\end{table*}

\section{Results and Analysis\label{exp-result}}

\noindent\textbf{Transductive \& Inductive Future Link Prediction.}
As shown in Table \ref{result-trans-induc}, (1) temporal methods surpass static ones, suggesting the importance of temporal properties in modeling temporal graphs;
(2) models assembled with FTM consistently outperform the originals on all benchmarks, demonstrating the effectiveness of FTM. For instance, on Wikipedia, FTM brings an average gain of \textbf{2.98} in AP under inductive setting. Meanwhile, TGN+FTM achieves new state-of-the-art performance on both Wikipedia and Reddit. The overall performance on this task indicates that FTM guides the learning of the evolution of temporal graphs and helps to generate more informative representations.

\subsection{Quantitative Analysis}
\label{Quantitative Analysis}
Given these overall performance improvements, we investigate how FTM's improvements are reflected in the learnt node representations. Because we have the gold label of node type in Wikipedia, we conduct a downstream task
of future link prediction, \textbf{node classification}, in two settings:
\emph{(i)} \textbf{Fine-tuning.} We fine-tune a MLP layer to classify nodes based on the learnt node embeddings. As the result in the second column of Table \ref{robust-attack-2} (attack intensity is 0) shows, FTM brings about 1\%\~{}4\% absolute gain in AUC to backbone methods, which reveals that models assembled with FTM generate more reasonable node embeddings. It also demonstrates the insights of our method in temporal graph representation learning;
\emph{(ii)} \textbf{Adversarial Attack.} 
The ability to resist Gaussian noise-perturbated examples is important because noisy data is inevitable under most circumstances \cite{DBLP:journals/corr/abs-2302-09328}. We add random Gaussian noise to the original data to generate adversarial examples for five times, and record the average performance of each model. The results are reported in the last six columns of Table \ref{robust-attack-2} (with attack intensity from 1\% to 50\%). The average gains that FTM brings to the baseline methods demonstrate that FTM can handle data noise (and maybe data biases) better, which is an important capability that guarantees the applicability of the proposed method.

\subsection{Qualitative Analysis\label{qualitative analysis}}

In this section, we examine our model's ability to generate more informative representations on the wikipedia dataset qualitatively.
As Figure \ref{node classify on wiki} shows, FTM helps to distinguish atypical users, whereas baselines are often misled; it reflects the potential of FTM in addressing data biases, since the data bias issues in data collected from platforms like Wikipedia are mainly caused by atypical users who often perform irregular/abnormal actions.
Moreover, we hypothesize that the evolution of user actions has short-term stationary features, because people's personality will not change rapidly. We take the most popular snapshot-based modeling method as the opponent to demonstrate that FTM makes it possible to capture short-term stationary features over time. First, we modify the neighborhood sampling strategy of the original TGAT to be snapshot-based, namely TGAT+Snapshot. Specifically, for each node we take its neighbors within an hour to form a temporal neighborhood. Then, we compute the cosine similarity of successive temporal node embeddings for TGAT+Snapshot and our TGAT+FTM respectively. As shown in Figure \ref{Fig.frame_snapshot_cos_sim}, the temporal node embeddings generated by TGAT+FTM show higher consistency. It demonstrates that TGAT+FTM learns more stable representations of users and we believe that the main reason lies in capturing short-term stationary features. Intuitively, this ability helps to stabilize the training process and capture the dynamics of user actions.

\subsection{Domain Generality} 
\label{domain generality}

Our reported results thus far demonstrate the effectiveness of FTM in improving the capability and robustness of temporal GNNs. In this section, we explore whether FTM could help improve the domain generality of baseline methods.
From the results shown in Table \ref{result-generalization-2}, we can observe that (1) these baseline methods suffer from severe domain generality issues, \emph{e.g.}, GraphSAGE trained on Reddit only get \textbf{46.89} in AP on Icews14; and (2) assembling FTM with these baseline methods greatly improves their domain generality, \emph{e.g.}, when applying models trained on Reddit to Icews14, FTM brings an average gain of \textbf{20.42} in AP to them. It illustrates the efficacy of FTM in deriving generalizable knowledge of graph evolution.
Furthermore, we test the capability of our method in handling domain gaps from a new perspective - we subsample user-action data from the wikipedia dataset with different time interval distribution and evaluate our method on it. The result shows that assembling FTM with baseline methods improves their AP by \textbf{1.5} in average, but is not listed here for space-saving issues.

\begin{table}[t]
\centering
\scriptsize
\setlength{\tabcolsep}{6.2pt}
\begin{tabular}{ccccc}
\toprule
\multirow{1}{*}{\textbf{Aggregation}} & \multicolumn{2}{c}{\textbf{AP on Wikipedia}} & \multirow{1}{*}{\textbf{Convergence}} & \multirow{1}{*}{\textbf{Parameter}} \\
\cline{2-3}
\textbf{Function} & \textbf{Transductive} & \textbf{Inductive} & \textbf{Time} & \textbf{Size} \\
\hline
1-layer MLP & \underline{97.68} & \underline{97.19} & $\textbf{8.5} \times \textbf{10}^\textbf{3}$ \textbf{s} & $\textbf{100}\textbf{\%}$ \\
2-layer MLP & 97.26 & 96.79 & $1.3 \times 10^{4}$ s & \underline{$105\%$}  \\
LSTM & 97.64 & 96.99 & $2.7 \times 10^{4}$ s & $112\%$ \\
Self-attention & \textbf{97.93} & \textbf{97.44} & \underline{$1.1 \times 10^{4}$ s} & $110\%$ \\
\bottomrule
\end{tabular}
\caption{\label{aggregate function}
Comparison of different aggregate functions in Timeline Aggregator module.}
\end{table}

\subsection{Case Studies}
In normal experiments, we set the number of model layers to be 2 and the length of frames to be 20 to form a node's temporal neighborhood. 
In this section, we record the performance of aforementioned methods under different neighborhood scales and data sizes. Note that the test data is the same as aforementioned experiments. 

In studying the influence of neighborhood scale, we separately let (the number of model layers, the length of frames) be $(1, 10)$, $(1, 20)$, $(2, 10)$, $(2, 20)$ to form a S-scale, M-scale, L-scale, XL-scale neighborhood respectively.
The results are provided in the left part of Table \ref{case-neighborhoodsize}. In all cases, models assembled with FTM outperform the originals. It illustrates that, even under low-resource settings, assembling FTM with backbone methods can enhance the capability, the robustness, and the domain generality of these models.

In studying the influence of data size, we sample $x$-percent of the training/validation set to form new training/validation sets. As the results in the right part of Table \ref{case-neighborhoodsize} illustrate, models assembled with FTM outperform the originals in most cases. It indicates that FTM is not totally data-driven, but superior in understanding the evolution of the temporal graph. This ability is of practical importance.

\subsection{Implementation \& Training Details}

\noindent\textbf{Hyper-parameters.} 
We do the chronological train-validation-test split with 70\%-15\%-15\% according to the timestamps of links. In the test set, we randomly sample 10$\%$ nodes as 'new nodes' for inductive tasks, and mask down all their links in the training set. Both the number of self-attention layers and the number of heads in each layer of the backbone network are 2. The length of timeline is chosen from [2, 3, 4] (we only report the best result). During training, we use Adam optimizer with learning rate 1e-4. The dimension of time encoding vectors is set to 172, which is same to the dimension of link feature vectors. We have conducted experiments to verify the effect of different aggregate functions in the Timeline Aggregator module. The result is shown in Table \ref{aggregate function} (timeline length is 2 and all experiments are conducted on a RTX 2080Ti GPU). Taking both the performance and efficiency into consideration, we decide to deploy a 1-layer MLP as the timeline aggregate function because it achieves comparable performance while having faster convergence rate and smaller parameter size than other aggregate functions. 
Readers can implement the self-attention mechanism for better performance. 

\section{Conclusion}
In this paper, we propose a simple but effective frame-level timeline modeling method for temporal graph representation learning, where the main contributions are made to the way that temporal neighborhoods are constructed and neighboring information is aggregated.
Technically, we break down a temporal sequence of graph-structured data into individual frames, and model the evolution of successive frames to mine deeper into the dynamics of nodes and links.
Experimental results demonstrate the effectiveness of FTM.
Meanwhile, 
our experiments empirically reveal that even state-of-the-art GNNs have critical weakness in modeling temporal graphs; but FTM helps to derive generalizable knowledge during training and thus greatly improves both the robustness and the domain generality of baseline methods, especially when there are outliers/noise in the data (\emph{cf.} Figure \ref{node classify on wiki}, Table \ref{robust-attack-2}), or the amount of data and computational resources are insufficient (\emph{cf.} Table \ref{case-neighborhoodsize}).
The efficacy of FTM may provide insights that could facilitate the design of more advanced representation learning methods on temporal graphs. 

\section{Acknowledgement}
This paper was partially supported by Shenzhen Science \& Technology Research Program (No: GXWD202012311658-\\07007-20200814115301001) and NSFC (No: 62176008)

\bibliography{aaai23}

\end{document}